\documentclass[lettersize,journal]{IEEEtran}
\pdfoutput=1
\usepackage{amsmath,amsfonts}
\usepackage{array}
\usepackage{algorithm}
\usepackage{algpseudocode}
\usepackage{textcomp}
\usepackage{stfloats}
\usepackage{url}
\usepackage{verbatim}
\usepackage{graphicx}
\usepackage{cite}
\usepackage{multirow}
\usepackage{hyperref}
\usepackage[caption=false,font=footnotesize,labelfont=rm,textfont=rm]{subfig}
\begin{document}

\title{Integrating Offline Pre-Training with Online Fine-Tuning: A Reinforcement Learning Approach for Social Robot  Navigation}

\author{Run~Su, Hao Fu, Shuai~Zhou, and Yingao~Fu
\thanks{*This work was supported in part by the National Natural Science Foundation of China under Grant 62303357 and Grant 62173262 and in part by the Hubei Provincial Natural Science Foundation of China under Grant 2023AFB109. \emph{(Corresponding author: Hao Fu.)}}

\thanks{R. Su, H. Fu, S. Zhou, and Y. Fu are with the School of Computer Science and Technology, Wuhan University of Science and Technology and also with the Hubei Province Key Laboratory of Intelligent Information Processing and Real-time Industrial System, Wuhan 430081, China (e-mail: fuhao@wust.edu.cn).}
}

\maketitle

\begin{abstract}
Offline reinforcement learning (RL) has emerged as a promising framework for addressing social robot navigation challenges. However, inherent uncertainties in pedestrian behavior and limited environmental interaction during training often lead to suboptimal exploration and distributional shifts between offline pre-training and online deployment. To overcome these limitations, this paper proposes a novel offline-to-online fine-tuning RL algorithm for social robot navigation  by integrating Return-to-Go (RTG) prediction into a causal transformer architecture. Our algorithm features a spatio-temporal fusion model designed to precisely estimate RTG values in real-time by jointly encoding temporal pedestrian motion patterns and spatial crowd dynamics. This RTG prediction framework mitigates distribution shift by aligning offline policy training with online environmental interactions. Furthermore, a hybrid offline-online experience sampling mechanism is built to stabilize policy updates during fine-tuning, ensuring balanced integration of pre-trained knowledge and real-time adaptation. Extensive experiments in simulated social navigation environments demonstrate that our method achieves a higher success rate and lower collision rate compared to state-of-the-art baselines. These results underscore the efficacy of our algorithm in enhancing navigation policy robustness and adaptability. This work paves the way for more reliable and adaptive robotic navigation systems in real-world applications.
\end{abstract}


\section{Introduction}

\IEEEPARstart{W}{ith} significant advancements in robotics and artificial intelligence, autonomous mobile robot navigation has garnered considerable attention. A primary challenge is developing a system that enables robots to move from a starting point to a desired target while effectively avoiding obstacles, especially in human-shared environments such as smart manufacturing, warehouses, and autonomous driving. This concept is referred to as socially-aware robot navigation. However, the complexity of pedestrian movement poses numerous challenges in designing effective  social robot navigation algorithms.

Socially aware robot navigation can be achieved through human-robot interaction, leveraging the inherent advantage of learning through trial and error.  Recently, significant efforts \cite{chen2019crowd,chen2020robot,everett2018motion} have been made in the field of socially aware robot navigation by incorporating deep learning techniques, such as Long Short-Term Memory (LSTM) and attention mechanisms.  These algorithms frame socially aware robot navigation as a Markov Decision Process (MDP), which is subsequently solved using value-based deep reinforcement learning (DRL).  Current robot navigation algorithms primarily focus on training policies in an online mode, learning navigation policies from raw sensory inputs, such as laser scans \cite{long2018towards}, images \cite{sathyamoorthy2020densecavoid}, or agent-level state representations \cite{chen2017decentralized}.

Current theses online reinforcement learning methods for social robot navigation necessitate frequent robot-pedestrian interactions within crowded settings. They rely on the iterative collection of extensive exploratory data to refine navigation policies. However, this training paradigm suffers from low sample efficiency, as it demands substantial volumes of interactive data to learn effective policies. Furthermore, the suboptimal policies characteristic of initial training phases can lead to unsafe exploration, presenting potential collision risks for both the robot and pedestrians.

In contrast, offline reinforcement learning, when applied to social navigation, leverages pre-existing datasets to optimize the navigation policy without requiring online interaction. This methodology significantly improves safety throughout the training process by eliminating risky exploratory actions. Nevertheless, the absence of online exploration and limited environmental interaction can impede the learning of a truly optimal navigation strategy.

To address the aforementioned challenges, the offline-to-online fine-tuning approach demonstrates significant potential. During its offline training, Return-to-Go (RTG) values are derived directly from empirical trajectory data by computing cumulative returns observed in the dataset. In contrast, during its online interaction, the use of fixed exploration RTG may exhibit discrepancies relative to dynamically generated returns in the real-world crowd scenario. Such misalignment can induce distribution shift problem, leading to aggressive or unsafe decision behavior.  This challenge is further compounded by uncertainties of pedestrian behavior. To mitigate this issue, we propose a OTOFRL (offline-to-online fine-tuning RL) algorithm. In particular, our algorithm trains a Return-to-Go prediction (RTGP) model for sequence modeling in the causal transformer, aiming to eliminate the distribution shift problem during the online fine-tuning phase caused by the complexity of pedestrian movements and the fixed RTG in the online setting. The key contributions of this study are summarized as follows:
\begin{itemize}

    \item To address the distribution shift issue during online fine-tuning, the OTOFRL algorithm is proposed through the establishment of an RTGP model based on a spatio-temporal fusion transformer and integrating sequence modeling with a causal transformer. By capturing the dynamic behavioral patterns of pedestrians in both temporal and spatial dimensions, the model can accurately predict long-term cumulative returns. This long-term return prediction enables the model to gain a more comprehensive understanding of environmental dynamics, thereby enhancing its adaptability to new data in human-robot interaction environments.
    \item To avoid the potential deviation issue, arising from synchronous updates between the robot navigation policy and the RTGP, this paper builds a hybrid offline-online sampling mechanism by incorporating a dual timescale update to manage the updates of these two components, so as to effectively reduces prediction variance and enhances the stability of policy adaptation.
    \item To ensure a seamless transition from offline pre-training to online fine-tuning, we propose a hybrid offline-online sampling method that combines hybrid offline-online experience replay with a prioritized sampling mechanism.

\end{itemize}

\subsection{Socially Aware Robot Navigation}

Socially Aware Robot Navigation refers to the movement of robots in spaces shared with pedestrians, where pedestrian behavior is often unpredictable and non-cooperative. Traditional reactive methods, such as  optimal reciprocal collision avoidance (ORCA)\cite{van2011reciprocal} and reciprocal velocity obstacle (RVO)\cite{van2008reciprocal}, specify interaction rules for a single step based on the current geometric configuration between robots and pedestrians. However, they fail to capture pedestrian behavior, leading to potentially unsafe movements. While trajectory-based methods can mitigate this issue, they inevitably encounter the "freezing" problem\cite{trautman2010unfreezing} in dense crowds. 

To address this issue, Chen et al. proposed a collision avoidance with DRL (CADRL) algorithm. To handle pedestrian behavior randomness, they extended it to Socially-Aware CADRL by introducing social norms \cite{chen2017socially}. However, these approaches require assumptions about specific motion models for neighboring agents over short time scales. To eliminate this need, Everett et al. used LSTM to extend CADRL, enabling it to accommodate varying pedestrian numbers. Additionally, self-attention mechanism has been employed to enhance DRL-based social navigation performance for improved crowd-robot interaction.

However, due to the limitations of online training, all these methods inevitably require frequent interactions with the environment to collect the data necessary for training the robot. Consequently, safety issue arises from collisions between navigating robots and pedestrians during exploration. Additionally, low sampling efficiency during pedestrian-robot interactions poses a significant challenge.

\subsection{Transformer for offline RL and Online Fine-tuning}

Recent advancements in RL have introduced a novel perspective that frames the offline RL problem as a context-conditioned sequence modeling task \cite{janner2021offline}, aligning RL with a supervised learning paradigm \cite{emmons2021rvs}. This approach shifts the focus from explicitly learning Q-functions or policy gradients to predicting action sequences conditioned on task specifications. For instance, Chen et al. \cite{radford2018improving} trained transformers as model-free, context-conditioned policies, while Janner et al. employed transformers for both policy and dynamics modeling, demonstrating that beam search could significantly enhance model-free performance. However, these studies primarily operate within the offline RL paradigm, analogous to fixed dataset training in natural language processing. Despite the promise of such methods, the prevailing paradigm in RL remains offline pre-training followed by online fine-tuning. Nair et al. \cite{ashvin2020accelerating} highlighted that applying offline or off-policy RL methods in this context often results in suboptimal performance, or even performance degradation, due to the accumulation of off-policy errors \cite{kumar2019stabilizing} and the excessive conservatism required in offline RL to mitigate overestimation in out-of-distribution states.

To address these challenges, various algorithms have been proposed. For example, Nair et al. developed an approach effective for both offline and online training regimes, while Kostrikov et al. \cite{kostrikov2021offline} introduced an expected implicit Q-learning algorithm that leverages behavior cloning to extract policies, thereby avoiding out-of-distribution actions and achieving robust online fine-tuning performance. Lee et al. \cite{lee2022offline} tackled the offline-to-online transition by balancing replay strategies and employing Q-function ensembles to preserve conservativeness during offline training. On the basis of the offline Decision Transformer (DT) \cite{chen2021decision}, Zheng et al. \cite{zheng2022online} enhanced the performance of the online fine-tuning phase by introducing an exploration mechanism and historical experience mixing in Online Decision Transformer (ODT). It is evident that excessive sampling of low-return experiences, such as those involving collisions between the robot and pedestrians, is detrimental to online fine-tuning. During the online phase, the hybrid offline-online sampling method is adopted to mitigate the issue of over-sampling low-reward experiences, thereby enhancing the model's ability to address challenges associated with online fine-tuning. By strategically focusing on high-reward and informative experiences, such as successfully navigating through dense crowds in a socially compliant manner, the sampling mechanism ensures more efficient learning and improved policy adaptation in  complex social navigation tasks.

\section{Methodology}
In this section, the Socially Aware Robot Navigation problem is
described. Then, the OTOFRL algorithm is presented.

\subsection{Problem Formulation} 
In addressing the navigation problem for mobile robots within the framework of RL, we formulate the navigation task as an MDP, represented by the tuple ($\mathcal{S}$, $\mathcal{A}$, $\mathcal{T}$, $\mathcal{R}$, $\gamma$). In this formulation, $\mathcal{S}$ denotes the state space of the agent, encompassing all possible configurations the robot may encounter in its environment. $\mathcal{A}$ represents the action space, which includes all feasible maneuvers the robot can execute at any given state. The transition probability $\mathcal{T}$ characterizes the likelihood of moving from one state to another, contingent upon the selected action. The reward function $\mathcal{R}$ quantifies the immediate feedback received by the agent following the execution of an action in a specific state, guiding the learning process. The discount factor $\gamma$ $\in$ (0,1] serves to prioritize immediate rewards over distant ones, thus influencing the agent’s decision-making strategy.

By systematically detailing these foundational components, we subsequently derive the RL formulation tailored for the social robot navigation problem, elucidating how these elements interrelate to enable effective navigation in dynamic and human-shared environments.
\subsubsection{State space} In a socially aware robot navigation environment, the state space at each time step consists of observable and unobservable states of the agents (robot and  pedestrians). The observable part includes velocity $\mathit{v}$ = [$\mathit{v_x, v_y}$], position $\mathit{p}$ = [$\mathit{p_x, p_y}$] and the radius $\bar{r}_{i}$ of the agent itself, while the unobservable part includes target position $\mathit{p_g}$ = [$\mathit{g_x, g_y}$], preferred velocity $\mathit{v_{pref}}$ and heading angle $\psi$. In this paper, a robot-centric frame defined in  \cite{chen2017decentralized}, is adopted to make the spatio-temporal state representation more general. Then, influence of the absolute position on decision-making is eliminated. The states of the robot and pedestrians after transformation are rewritten by
\begin{equation}
\mathit{s}^r_t =[\mathit{d_g},\mathit{v_x},\mathit{v_y},\mathit{v_{pref}},\bar{r}_{0},\psi],
\end{equation}
\begin{equation}
\mathit{s}^i_t =[\widetilde{p}^i_x,\widetilde{p}^i_y,\widetilde{v}^i_x,\widetilde{v}^i_y, \bar{r}_{i}, d_{i}, \bar{r}_{i}+\bar{r}_{0}], i=1,2,\ldots,m
\end{equation}
\begin{equation}
\mathit{s}_t =[\mathit{s}^0_t,\mathit{s}^1_t,...,\mathit{s}^m_t],
\end{equation}
where $\mathit{s}^r_t$ and $\mathit{s}^i_t$ are the states of the robot and the $i$-th pedestrian at time $t$, $\mathit{d_g}$ = ${\parallel\mathit{p_g}-\mathit{p}\parallel}_2$ is the robot’s distance to the goal, $d_{i}$ = ${\parallel\mathit{p}-\mathit{p_i}\parallel}_2$ is the robot’s distance to the pedestrian $i$.

\subsubsection{Action space} This paper employs continuous actions to control the movement of the robot. specifically, the robot action can be expressed at time step $t$ as:
\begin{equation} \label{reward}
\mathit{a_t} = [\mathit{v_x},\mathit{v_y}],
\end{equation}

\subsubsection{Reward function} Ensuring safe robot navigation in crowds requires the robot to adhere to human social norms while efficiently reaching its destination.  Its reward function should be formulated to encourage successful navigation while penalizing collisions and overly close encounters with pedestrians.  It can balance efficiency, safety, and social compliance, guiding the robot to generate smooth and socially aware trajectories. Consequently, the reward function is designed as
\begin{equation}
r_{t}(s_t, a_t) = 
\begin{cases}
    -0.25, & \text{if } d_{\min}^{t} \leq 0 \\
    d_{\min}^{t} - 0.2, & \text{else if } d_{\min}^{t} < 0.2 \\
    2, & \text{else if } d_{g}^{t} \leq  \bar{r}_{0} \\
    0, & \text{otherwise}
\end{cases}
\end{equation}
where $d_{min}^t$ is the distance between the robot and the nearest pedestrian, and $d^t_g$ is the distance between the goal and the robot at time $t$. 

\subsection{OTOFRL for Socially Aware Robot Navigation}
In the context of socially aware robot navigation, the transition from  offline-to-online reinforcement learning is particularly susceptible to the issue of distribution shift. Specifically, offline reinforcement learning relies on pre-collected static datasets for policy optimization, while online fine-tuning requires real-time policy adjustments in social navigation environments. Due to the high complexity of pedestrian movements in social navigation scenarios, the state-action distribution in offline datasets often fails to fully cover the true distribution in the human-robot interaction environment. This discrepancy leads to significant distribution shifts when the policy is deployed, which can impair the policy's generalization capability and result in performance degradation or even safety risks. Therefore, effectively mitigating distribution shift during the transition from offline to online fine-tuning has become a critical challenge in enhancing the robustness and adaptability of social robot navigation systems.

\begin{figure*}[t]
\centering
\includegraphics[scale=0.5, trim={50mm 40mm 50mm 40mm}]{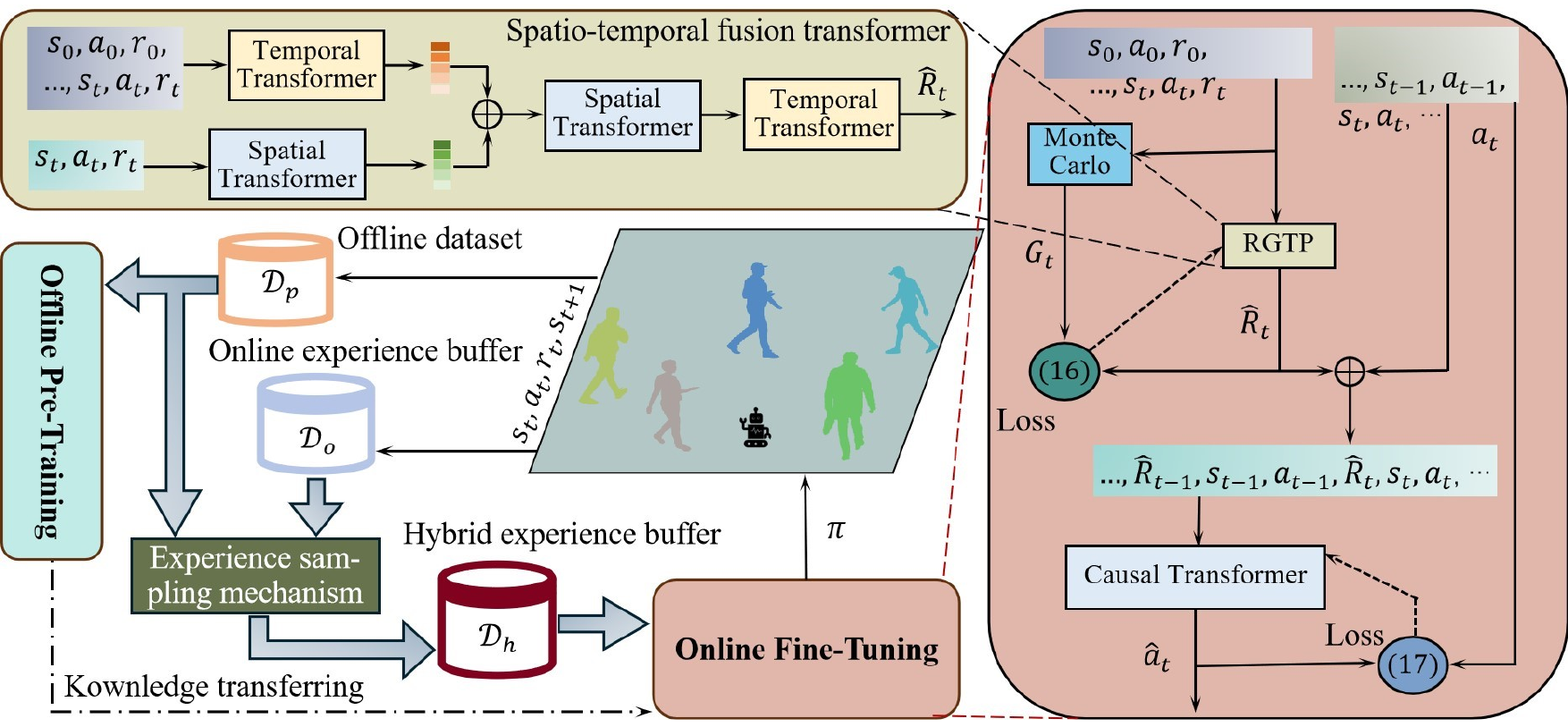}
	\caption{The OTOFRL architecture employs online fine-tuning to adapt offline DT and RTGP models through knowledge transfer from offline learning. A spatio-temporal fusion transformer predicts the RTG, which is then used as a token in the online DT. Both the online DT and RTGP models are subsequently updated using a hybrid offline-online sampling mechanism.}
	\label{Network Architecture}
\end{figure*}

To better address the distribution shift problem, we propose a RTGP model constructed using a spatio-temporal fusion transformer. The dynamic features of pedestrians in both temporal and spatial dimensions are captured for more accurate predictions of their future behaviors, leading to more reliable RTG estimates for robot navigation policies. Specifically, the multi-head self-attention mechanism of the transformer is leveraged to effectively integrate spatio-temporal information from historical pedestrian trajectories, extracting key features of their movement patterns. This fusion of spatio-temporal features not only enhances the model's ability to predict short-term pedestrian behaviors but also improves its inference accuracy for long-term trends. The detailed architecture and implementation of the model are illustrated in Figure~\ref{Network Architecture}, further demonstrating its advantages in handling complex social navigation scenarios.

To effectively capture the spatio-temporal dynamics of global states, we represent the trajectory data in the dataset as a spatio-temporal sequence. We employ a spatial state encoder and a temporal state encoder to capture high-level spatio-temporal features, which are then used to train the RTGP model. This approach allows the model to better leverage the spatio-temporal information while ensuring efficient generalization during the online fine-tuning phase. 

We define the spatial sequence 
$E_{s}=[s_{t}, a_{t}, r_{t}]$ as the input to the spatial transformer.
$E_{s}$ is embedded into a higher-dimensional space for preliminary feature extraction, with the extracted feature $\mathit{f}$ represented as:
\begin{equation}
\mathit{f} = \mathit{f_p}(E_{s};W_p),
\end{equation}
where $\mathit{f}_p$ is a fully connected layer with rectifed linear unit (ReLU) activation, and $W_p$ is the weight of parameters. The global spatial state encoder is used to capture the positional relationships between different pedestrians and the robot. This encoder highlights the importance of different pedestrians to the robot through a spatial multi-head self-attention layer. A feedforward neural network (FNN) then maps the spatial relationships into a high-dimensional feature space. It takes the initially extracted features $\mathit{f}$ as input and outputs enhanced features with spatial dependencies. The following are the equations for the global spatial encoder:
\begin{equation}
\mathit{Q}_s = \mathit{f}_{qs}(\mathit{f};W_{qs}),
\end{equation}
\begin{equation}
\mathit{K}_s = \mathit{f}_{ks}(\mathit{f};W_{ks}),
\end{equation}
\begin{equation}
\mathit{V}_s = \mathit{f}_{vs}(\mathit{f};W_{vs}),
\end{equation}
where $\mathit{f}_{qs}$, $\mathit{f}_{ks}$ and $\mathit{f}_{vs}$ are fully connected layer with ReLU activation, where $W_{qs}$, $W_{ks}$ and $W_{vs}$ represent the weight parameters, and $Q_s$, $K_s$ and $V_s$ denote the query, key and value vectors, respectively. Spatial dependencies are captured using the multi-head self-attention mechanism, which summarizes the attention scores of each pedestrian relative to the robot.
\begin{equation}
\mathit{Att}_i(\mathit{Q}_s,\mathit{K}_s,\mathit{V}_s) = \mathit{softmax}(\frac{Q_sK^T_s}{\sqrt{d_k}})V_s,
\end{equation}
\begin{equation}
\mathit{head}_i = \mathit{Att}_i(\mathit{Q}_s,\mathit{K}_s,\mathit{V}_s),
\end{equation}
\begin{equation}
\mathit{spatial-MSA}(\mathit{Q}_s,\mathit{K}_s,\mathit{V}_s) = \mathit{f}_o([\mathit{head}_i]^h_i),
\end{equation}
where $\mathit{Att}_i(\mathit{Q}_s,\mathit{K}_s,\mathit{V}_s)$ is a self-attention head, and $\mathit{f}_o$ is the fully connected layer that merges $\mathit{h}$ heads. $\mathit{d}_k$ denotes the dimensionality of the query and key vectors, The output of the multi-head attention layer is fed into the FNN through a residual connection and a normalization layer: 
\begin{equation}
f^{\prime}_{s} = \mathit{Spatial-MSA}(\mathit{Q}_s,\mathit{K}_s,\mathit{V}_s)+\mathit{f},
\end{equation}
\begin{equation}
f_{s} = FNN(LN(f^{\prime}_{s}))+f^{\prime}_{s},
\end{equation}
where FNN refers to a two-layer fully connected neural network with ReLU activation.

Similarly, the temporal sequence $E_{p}=[s_{0}, a_{0}, r_{0}, s_{1}, a_{1}, r_{1}, \dots, s_{t}, a_{t}, r_{t}]$  is processed by a causal transformer. The spatial and temporal transformers, operating in parallel, independently extract respective spatial and temporal features. These features are subsequently integrated via a fully connected layer, producing a new set of spatio-temporal encodings. To further capture spatio-temporal interactions, the features are processed sequentially by an additional spatial transformer followed by a temporal transformer. The spatial transformer models spatial relationships conditioned on the temporal information, while the temporal transformer refines the resulting spatial embeddings with temporal attention. This architecture enhances the model's capacity for spatio-temporal representation, leading to significant improvements in the RTGP.

The interaction between humans and the robot over time plays a crucial role in the robot's decision-making. The local temporal state encoder captures the temporal dynamics of each pedestrian through a multi-head self-attention layer along the time dimension. Similarly, the FNN is used to map the temporal evolution cues into a high-dimensional feature space.

To emphasize the temporal dimension, $[s_{t}, a_{t}, r_{t}]$ is also fed into a local temporal state encoder for initial feature extraction, with its formulation mirroring that of the local spatial state encoder. The spatial and temporal features are then merged through a fully connected layer to create a new set of features encoded in both space and time. To further model the spatio-temporal interactions in the feature space, the new features are input into a spatial transformer that simulates spatial interactions using temporal information. This is followed by inputting the features into a temporal transformer that enhances the spatial embeddings while increasing temporal attention. Finally, the resulting spatio-temporally enriched features are fed into a separate network to predict the RTG. Then, a RTG predictor is given by
\begin{equation}
\hat{R} = \mathit{f}_R(\mathit{Average(\mathit{f}_{st}),\mathit{s}_t};\mathit{W}_R),
\end{equation}
where $\mathit{f}_R$ represents a two-layer fully connected network with ReLU activation, and $\mathit{W}_R$ denotes the weight matrix of the network. $\mathit{f}_{st}$ refers to the enhanced features generated by the spatio-temporal transformer within the network. The network is trained using the Monte Carlo reinforcement learning method\cite{sutton1999reinforcement}, with the loss function defined as follows:
\begin{equation} \label{loss_RGT}
L_R = E_{(s,a,G)\sim\mathcal{D}}(G_{t} - \hat{R}(s_{t},a_{t}))^2,
\end{equation}
where $G_{t}$ is the Monte Carlo return, $\gamma$ is the reward discount factor, $\mathcal{D}$ is an experience replay, and $r_t$ is the actual reward obtained by the robot at time step $t$.

Instead of the fixed RTG in online DT \cite{zheng2022online}, the RTG predictor is integrated into the sequence modeling process of the causal transformer: $(\hat{R}_1,s_1,a_1,...,\hat{R}_t,s_t,a_t)$. Causal transformer learns a deterministic policy $\pi(a_t|s_{-K,t},\hat{R}_{-K,t})$, where $s_{-K,t}$ is shorthand for the sequence of $K$ past states. The policy is trained to predict action tokens under the following loss function:

\begin{equation} \label{loss_off}
L_{NT} = E_{(s,a,\hat{R})\sim\mathcal{D}}[\frac{1}{K}\textstyle\sum_{t=1}^{K}(a_k - \pi(s_{-K,t},\hat{R}_{-K,t}))^2.
\end{equation}

The combination of the online DT and the RTGP can mitigate discrepancies relative to dynamically generated returns, arising from the fixed exploration RTG. This further settles the distribution shift problem, reducing aggressive or unsafe decision behavior. Nevertheless, the transition from offline to online fine-tuning also poses several critical challenges that can hinder RL performance. First, offline RL models often exhibit over-conservatism due to the need to prevent overestimation of out-of-distribution actions, which can severely limit exploration during the online phase.  Second, the reliance on static datasets in offline pre-training restricts the model's ability to adapt dynamically to novel states or trajectories encountered online, leading to suboptimal generalization.  Third, during online fine-tuning, the accumulation of off-policy errors may degrade policy performance, especially when the model encounters scenarios not represented in the offline dataset.  Finally, synchronous updates between the online DT  and the RTGP model inevitably result in the potential deviation issue.

To mitigate the challenges inherent in transitioning from offline pre-training to online fine-tuning, we devise a hybrid
offline-online sampling mechanism combining a priority sampling strategy from a hybrid experience replay and a dual timescale update rule. Specifically, a hybrid experience replay buffer $\mathcal{D}_{h}$ is built by blending newly acquired online experiences in the online experience replay buffer $\mathcal{D}_{o}$ into pre-collected offline dataset $\mathcal{D}_{p}$.
Then, a priority sampling strategy is introduced by assigning more important experiences that are deemed more critical for online fine-tuning from the hybrid experience replay, such as those associated with novel, uncertain, or high-risk interactions. Furthermore, a dual-timescale update rule is employed: the online fine-tuning process, governed by loss function (\ref{loss_off}), is updated on a slow timescale using trajectories sampled from $\mathcal{D}_{h}$, while the RTGP model, with loss function (\ref{loss_RGT}), is updated on a fast timescale using individual transitions sampled from $\mathcal{D}_{h}$.

\textbf{Remark 1:} By mediating between the offline dataset $\mathcal{D}_{o}$ and the online experience replay buffer $\mathcal{D}_{p}$, our hybrid offline-online sampling mechanism seamless integration of both datasets. This enables the model to preserve the stability of offline pre-training while adapting to online interactions. The accompanying dual-timescale update rule further ensures stable policy adaptation by reducing prediction variance during the transition from pre-training to fine-tuning.

In conclusion, the proposed OTOFRL algorithm leverages a combination of the priority sampling from hybrid experience replay and a dual timescale update to optimize the transition from offline to online learning. This can adapt to real-world dynamics with greater efficiency and safety, ultimately resulting in a more robust and reliable robotic navigation system. In detail, the proposed OTOFRL algorithm is illustrated in Algorithm ~\ref{alg2}.

\begin{algorithm}[t]
    \caption{OTOFRL}
    \label{alg2}
    \begin{algorithmic}[1]
        \State \textbf{Input} Offline dataset $\mathcal{D}_p$, episode number $N$, RTGP parameter $\phi$, DT parameter $\theta$, hybrid replay buffer $\mathcal{D}_{h}$, iteration number $I$, episode number $I$, context length $K$, batch size $B$
        \State \textbf{Initialize} Hybrid replay buffer $\mathcal{D}_{h}$, online experience replay buffer $\mathcal{D}_{o}$,  $\phi$, $\theta$.
        \While{Convergence} \Comment{Offline pre-training}
            \State Sample a random mini-batch trajectories from $\mathcal{D}_p$ 
            \State Compute action sequences $\pi(s_{-K,t},R_{-K,t})$ and prediction RTG $\hat{R}_{t}$
            \State Update parameter $\theta$ in offline DT
            \State Compute Monte Carlo turn $G_{t}$
            \State Update parameter $\phi$ in the RTGP via (\ref{loss_RGT})
        \EndWhile
        \For{$episode=1,\ldots,N$}  \Comment{Online fine-tuning}
            \While{robot not reach goal, collide or timeout}
                \State Calculate prediction RTG $\hat{R}_{t}$
                \State Feed trajectory sequence into online DT to get prediction action $a_{t}$
                \State Execute action $a_{t}$ and obtain new state $s_{t+1}$ and reward $r_{t}$  
            \EndWhile
            \State  Assimilate trajectory into $\mathcal{D}_{o}$
            \State Obtain hybrid experience replay $\mathcal{D}_{h}$
            \State Sample a random mini-batch trajectories $\tau$ from $\mathcal{D}_{h}$ via sampling mechanism
            \For{each sampled trajectory $\tau$}
                \State Compute action $\pi(s_{-K,t},\hat{R}_{-K,t})$ and prediction RTG $\hat{R}_{t}$
                \State Update parameter $\phi$ via (\ref{loss_RGT}) under a fast time scale
            \EndFor
            \State Compute Monte Carlo Turn $G_{t}$
            \State Update parameter $\theta$ via (\ref{loss_off}) under a slow time scale 
        \EndFor
    \end{algorithmic} 
\end{algorithm}

\section{EXPERIMENTS}

\subsection{Simulation Setup}

1) $\mathit{Simulation\ environment}$: In each episode, the robot starts from the initial position (0, -4) and aims to reach the goal at (0, 4). The circle crossing environment is used for both training and testing. 5 pedestrians begin at positions located on a circle with a 4-meter radius, and their target positions are on the opposite side of the same circle. To reflect the unpredictability of real-world environments, random perturbations are introduced to both the pedestrians' initial and target positions. Finally, once a pedestrian reaches their target, they are assigned a new randomly generated destination.

2) $\mathit{Creating\ Datasets}$: Our dataset is constructed within this simulated environment, where the robot is set to be invisible. In this setting, the robot is prone to colliding with pedestrians. To address this issue, a safety space is incorporated into the robot's policy to ensure it can successfully avoid pedestrians to some extent. The robot's safety space is set to 0.02 to demonstrate the effectiveness of offline pre-training and online fine-tuning.

We collected data from the simulated environment and created a dataset. Table 1 details this dataset using five metrics: "Success," "Collision," "Time," "Reward," and "Capacity." These metrics describe the trajectory's success rate, collision rate, average navigation time, average cumulative returns across all trajectories, and the dataset's capacity, respectively.

3) $\mathit{Baseline}$: We compare against six state-of-the-art algorithms. ORCA \cite{van2011reciprocal} is used as the reactive method baseline; CQL \cite{kostrikov2021offline} and DT \cite{chen2021decision} are the offline reinforcement learning baseline. ODT\cite{zheng2022online} is the offline-to-online reinforcement learning baseline. LSTM-RL \cite{everett2018motion}, SARL \cite{chen2019crowd}, and DS-RNN \cite{liu2021decentralized} are used as baselines for traditional DRL-based robotic crowd navigation methods.

4)$\mathit{Training\ Settings}$:All the aforementioned algorithms are trained using the same set of environmental hyperparameters. 

In our algorithm, each individual network uses the LAMB optimizer \cite{you2019large}. The three fully connected networks have dimensions of (65, 128), (65, 128) and (256, 1), respectively. Each network is also equipped with layer normalization and ReLU activation functions. The key hyperparameter values are listed in Table~\ref{para}.

\begin{table}[htbp]
	\setlength{\tabcolsep}{7pt}
	\caption{Hyperparameter}
	\begin{center}
		\begin{tabular}{lclc}
			\hline
			Parameter  &  Value  & Parameter  &  Value \\
			\hline
			Learning rate & 5$\times10^{-4}$ & Batch size  &  256 \\
			Replay memory size $\mathcal{D}$ & $10^{5}$ & $v_{\max}$ & 1.0 m/s \\
			Maximum episode  & $10^{4}$  &  Maximum time  &  25 s   \\
			Discount factor $\gamma$  &  0.99  & &  \\
			\hline
		\end{tabular}
		\label{para}
	\end{center}
\end{table}

In the implementation of ORCA, the robot's safety space is set to 0.02, consistent with the policy in the dataset. CQL, DT and ODT use the same dataset and training parameters as our algorithm. LSTM-RL, SARL, and DS-RNN follow the same reward function defined in Equation 5, with their network architectures and training settings remaining consistent with the original papers.

\subsection{Quantitative Evaluation}

During testing, all methods are evaluated across 500 test environments. In this experiment, the robot is set to be invisible, requiring it to avoid collisions to reach the target successfully. The table summarizes the "Success Rate," "Collision Rate," "Average Navigation Time," and "Average Reward" over the 500 test environments. Additionally, to assess performance related to sampling efficiency, we include the "Sampling Efficiency" metric, which quantifies the efficiency of all methods. "Sampling Efficiency" is defined as:
\begin{equation}
\eta =  \frac{r}{U},
\end{equation}
where $\eta$ represents the sampling efficiency, $r$ represents the average reward, and $U$ represents the sample size, The experimental results are summarized in Table~\ref{ex}.

As shown in Table~\ref{ex}, due to the minimum safety distance being set to 0.02, the baseline algorithm ORCA has a relatively low success rate. Comparing the two offline learning algorithms, CQL, DT, and our method all achieve higher success rates and average rewards than the ORCA baseline, indicating that both can learn better policies from suboptimal strategies. Compared to CQL, our algorithm outperforms across all metrics. This is because CQL focuses more on short-term conservatism and lacks long-term path planning, whereas our method emphasizes long-term path optimization. As a result, after training, our method demonstrates a more optimal policy. Compared to DT, our success rate improves by approximately 18\%, navigation time is significantly reduced, and the average reward increases accordingly. This suggests that our algorithm effectively enhances policy feasibility during the online phase. Compared to ODT, another offline-to-online learning algorithm, our method performs better on all metrics except for navigation time, where ODT has a slight advantage. This is because ODT’s random exploration in the online phase can yield shorter navigation paths but also introduces higher risks. Compared to the other two traditional online learning baselines,  LSTM-RL, and DSRNN, our algorithm outperforms them across all metrics. Additionally, our algorithm surpasses SARL in all but the navigation time metric. This is because, during the online learning phase, we employed priority sampling, prioritizing trajectories with higher cumulative returns upon success. Consequently, our algorithm favors a strategy focused on success and maximizing cumulative returns.

From the last column of the table, it can be seen that our algorithm's sampling efficiency is nearly 31\% higher than that of the best-performing traditional DRL baseline, SARL. This is because traditional online learning DRL baselines typically require frequent interactions with the environment to acquire diverse learning experiences. Additionally, our algorithm outperforms ODT in sampling efficiency by nearly 14\%. This is because ODT adopts a random exploration strategy during the online phase, requiring more experiences for learning. Finally, our algorithm achieves higher sampling efficiency than both offline algorithms, CQL and DT. These results indicate that combining causal transformer with the RTGP model effectively enhances the robot’s safe exploration ability during online training, thereby improving sampling efficiency.

\begin{table}[h]
\caption{Quantitative results of all methods}
\centering
\begin{tabular}{c c c c c c}
\hline
\textbf{Methods} & \textbf{Success} & \textbf{Collision} & \textbf{Time} & \textbf{Reward} & \textbf{Efficiency} \\
ORCA & 61.2\% & 38.8\% & 11.75 & 0.3405 &  \\
LSTM-RL & 97.0\% & 3.0\% & 11.94 & 0.7157 & 0.110\\
SARL & 99.0\% & 1.0\% & 10.13 & 0.8771 & 0.136\\
DS-RNN & 96.0\% & 4.0\% & 12.00 & 0.8018 & 0.101\\
CQL & 63.0\% & 36.0\% & 13.39 & 0.5181 & 0.104\\
DT & 81.0\% & 19.0\% & 11.55 & 0.6647 & 0.133\\
ODT & 95.0\% & 5.0\% & 10.43 & 0.8309 & 0.156\\
OTOFRL & 99.6\% & 0.4\% & 11.29 & 0.9811 & 0.178\\
\hline
\label{ex}
\end{tabular}
\end{table}

\begin{figure*}[t]
\centering
\subfloat[ORCA]{
\includegraphics[trim=60 180 94 80, clip, scale=0.25]{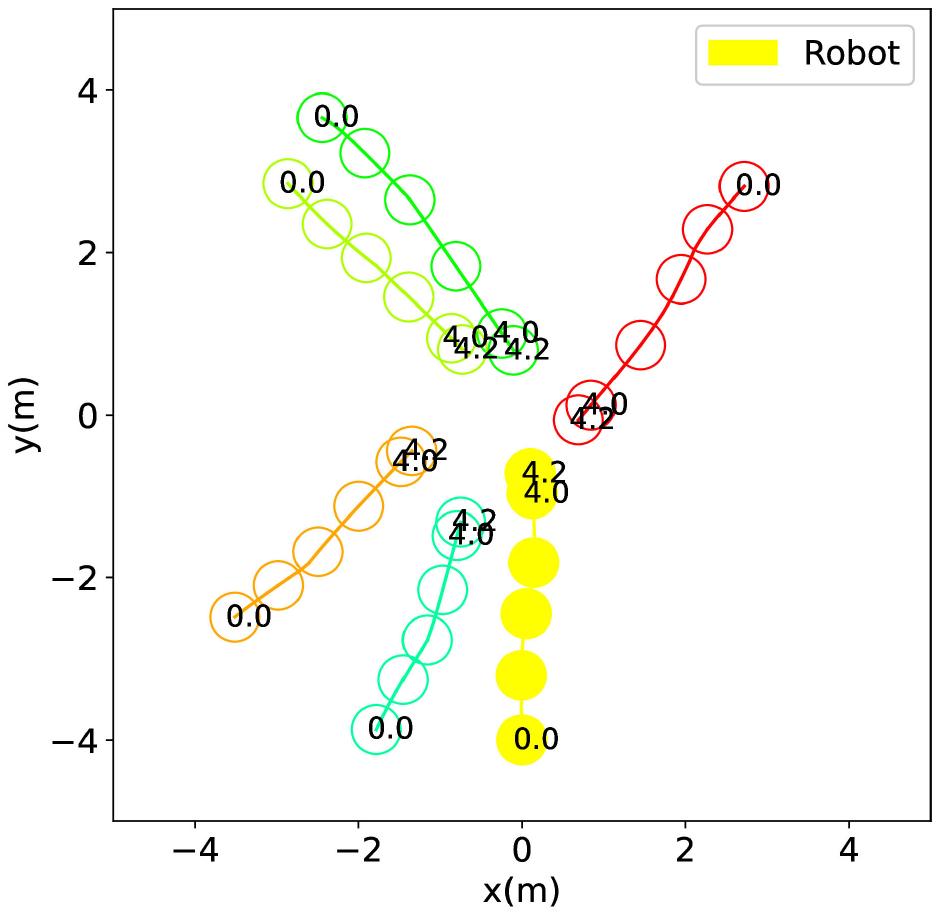}
\label{Testing1a}} 
\quad
\subfloat[LSTM-RL]{
\includegraphics[trim=60 180 94 80, clip, scale=0.25]{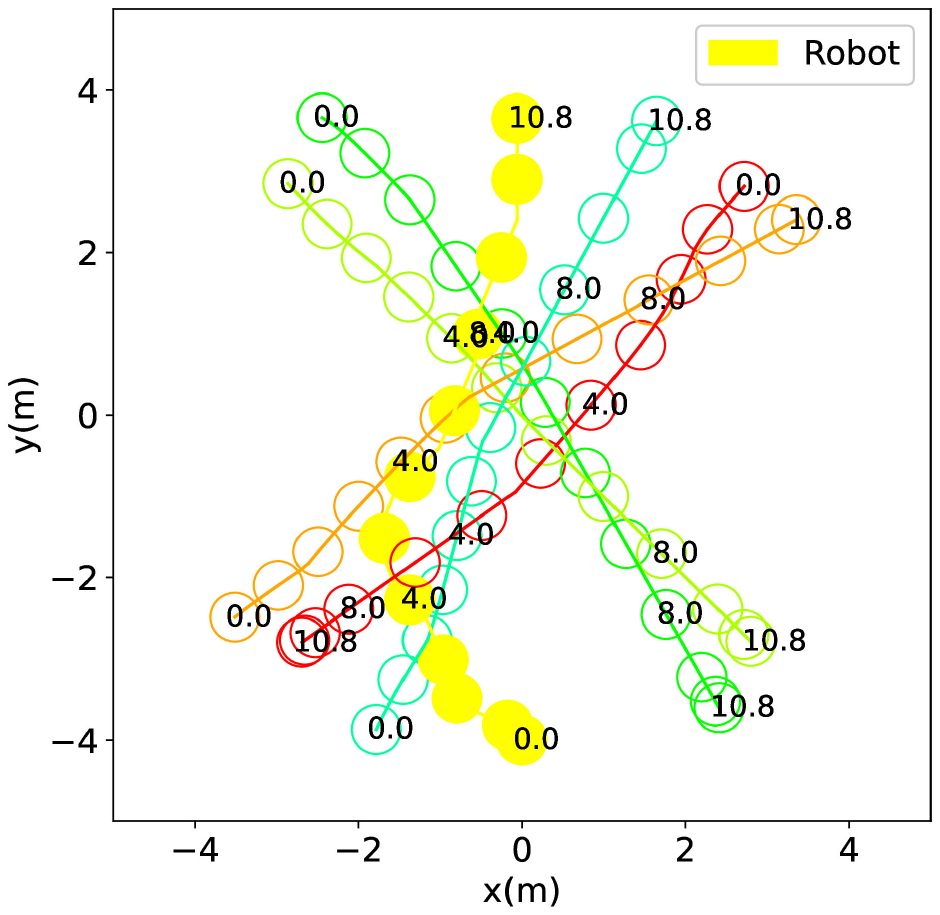}
\label{Testing1b}
}
\quad
\subfloat[SARL]{
\includegraphics[trim=60 180 94 80, clip, scale=0.25]{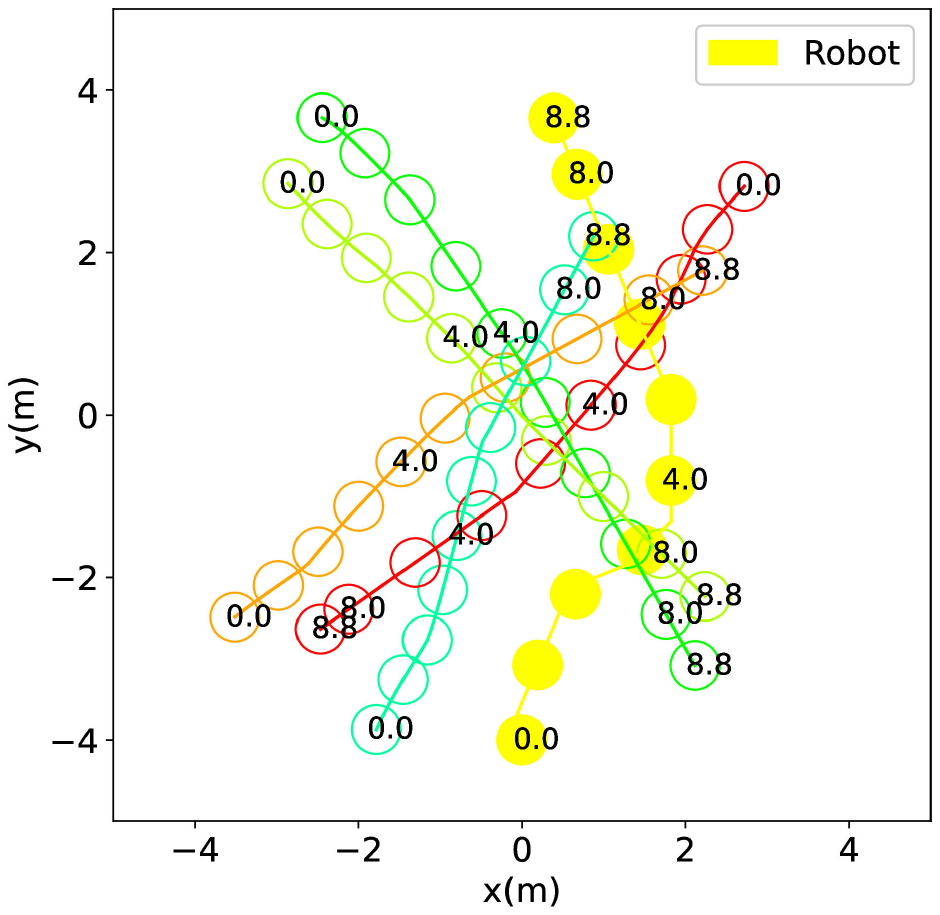}
\label{Testing1c}
}
\quad
\subfloat[DSRNN]{
\includegraphics[trim=60 180 94 80, clip, scale=0.25]{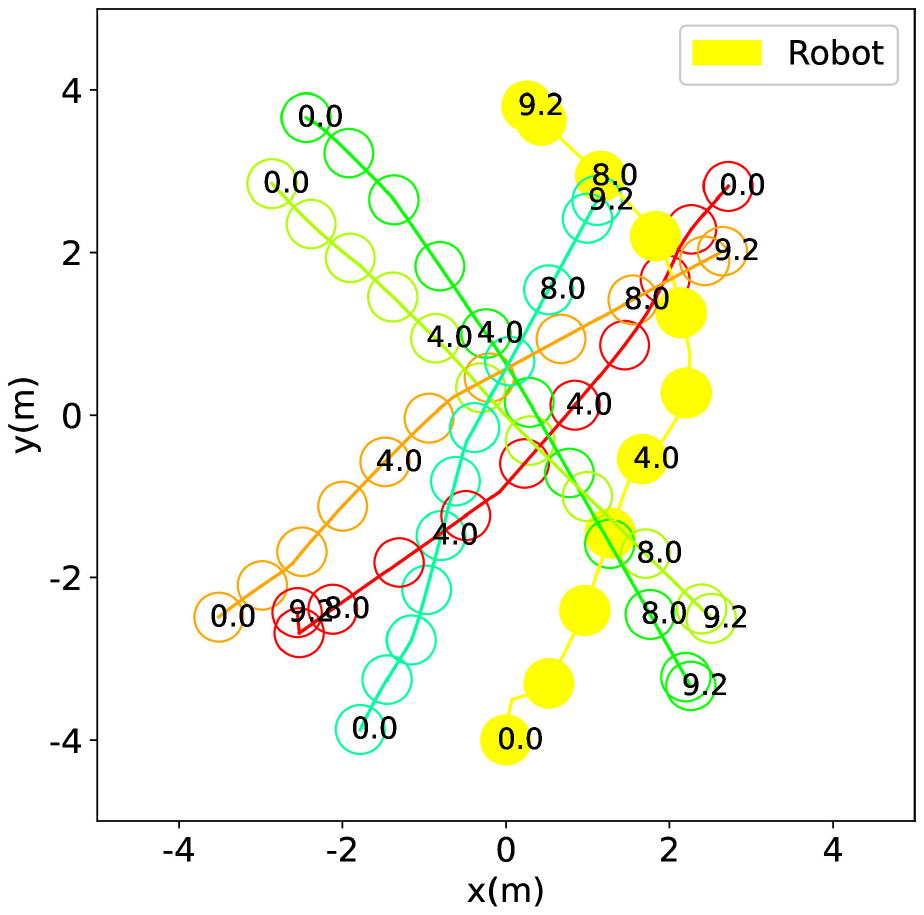}
\label{Testing1d}
}
\quad
\subfloat[CQL]{
\includegraphics[trim=60 180 60 50, clip, scale=0.25]{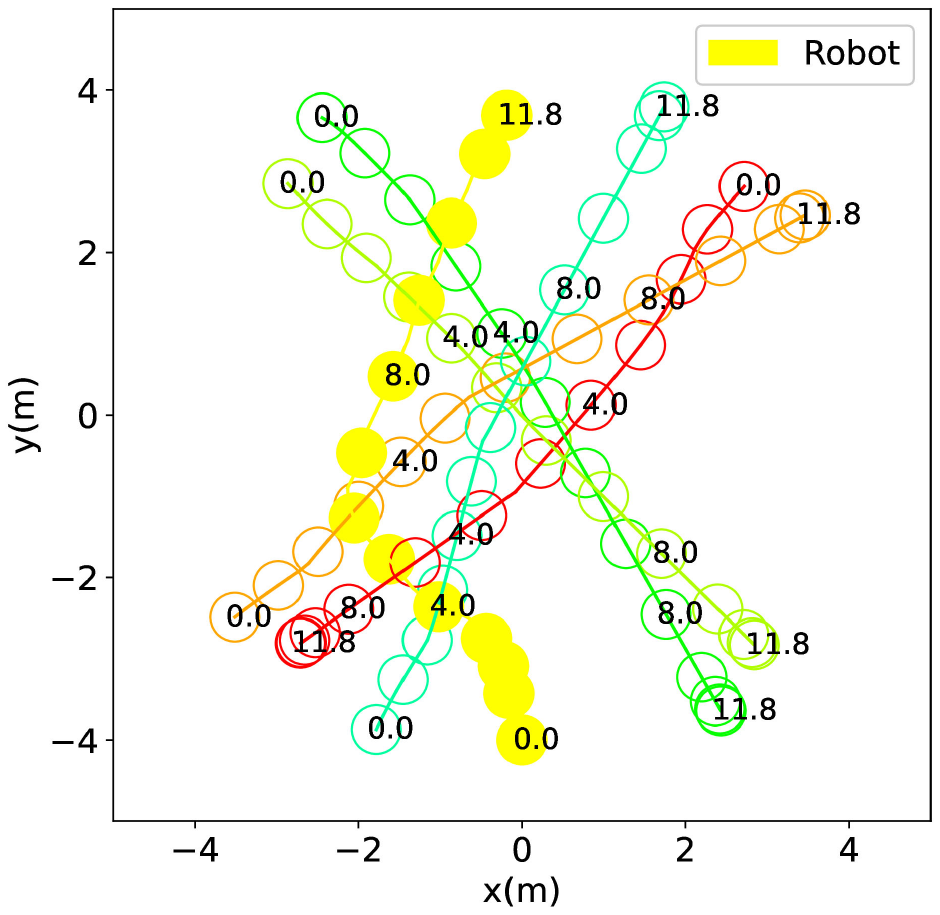}
\label{Testing1e}
}
\quad
\subfloat[DT]{
\includegraphics[trim=98 180 94 80 clip, scale=0.25]{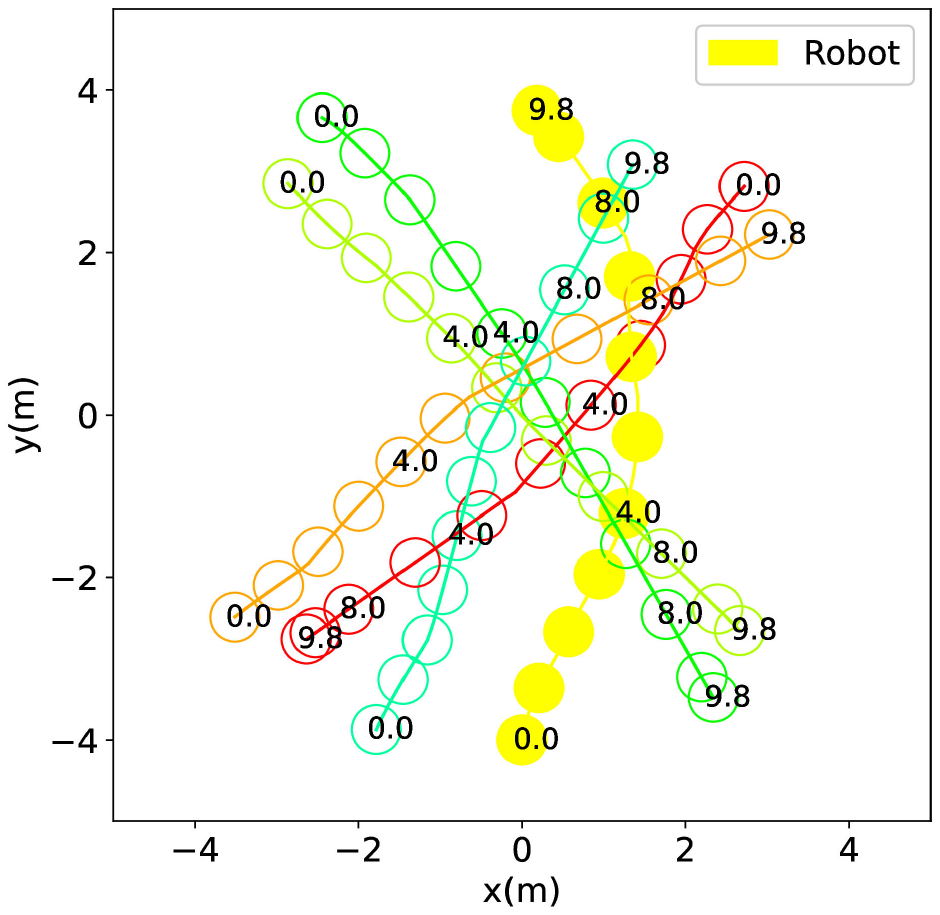}
\label{Testing1f}
}
\quad
\subfloat[ODT]{
\includegraphics[trim=60 180 94 80, clip, scale=0.25]{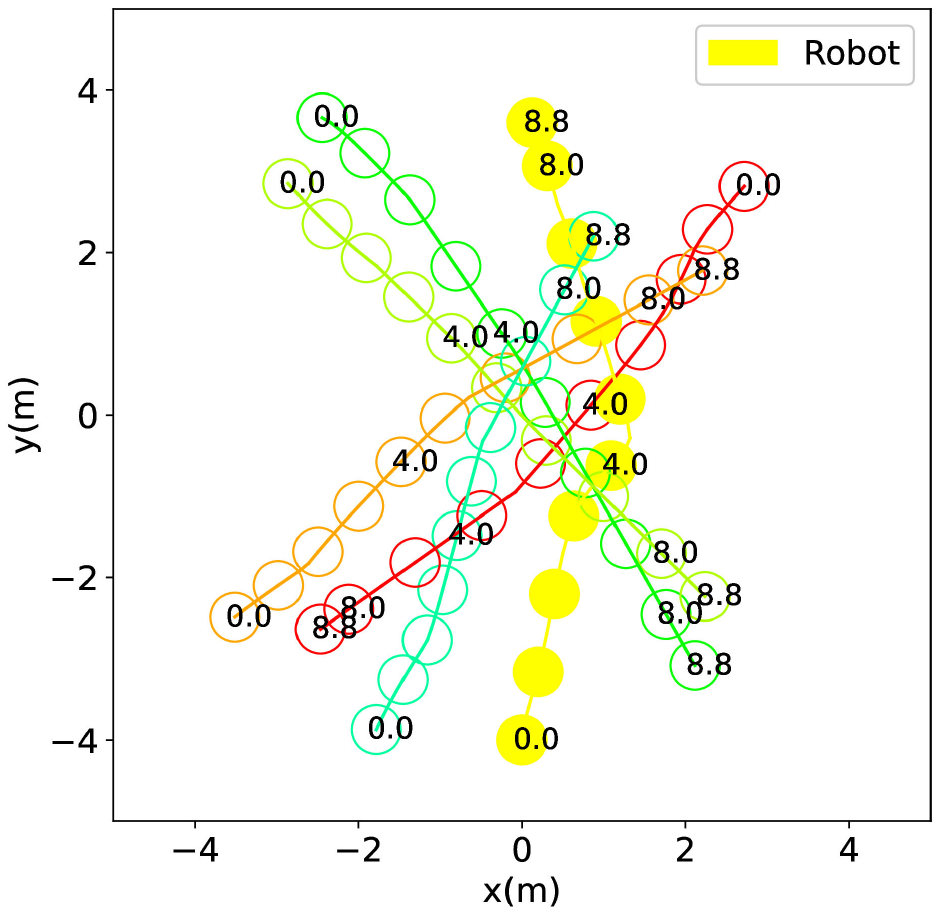}
\label{Testing1g}
}
\quad
\subfloat[OTOFRL]{
\includegraphics[trim=60 180 94 80, clip, scale=0.25]{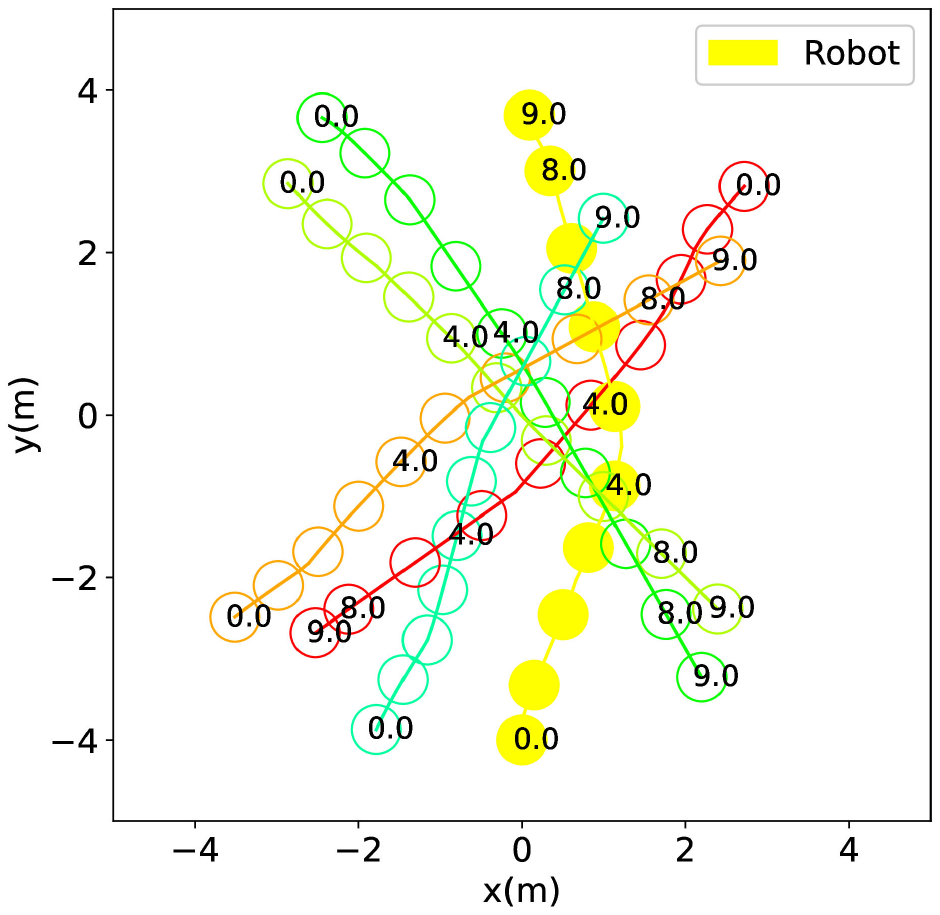}
\label{Testing1h}
}
\caption{Robot trajectory comparison of the different methods in identical social formation navigation test scenarios.}
\label{Testing2}
\end{figure*}

\subsection{Qualitative Evaluation}
To rigorously evaluate our algorithm through qualitative analysis, we conduct a visual examination of the robot’s behavior in randomly generated, dynamically populated crowd scenarios, comparing global trajectories produced by various navigation methods.
The ORCA algorithm demonstrates a tendency to remain close to the crowd, frequently resulting in navigation failures. The LSTM-RL approach exhibits hesitancy at the start of navigation, consequently prolonging the overall travel time. While the SARL method achieves the shortest navigation duration, it exclusively relies on pedestrian state information, yielding unnatural trajectories and a lack of deceleration in densely populated areas, thus compromising safety. In contrast, the DSRNN often selects detours, even when the initial distance from the crowd is ample, which also extends the navigation time. The CQL algorithm, focused on short-term conservatism, tends to maintain greater distance from crowds during navigation, further increasing travel time. DT, by emphasizing long-term planning, generates comparatively improved trajectories; however, limited exploration restricts its ability to identify optimal paths. ODT achieves more natural trajectories than SARL due to its incorporation of long-term planning and exploration. However, its lack of deceleration in densely populated areas compromises safety.
In comparison, our proposed NaviTune-Transforme algorithm refines the trajectory further, enhancing the route efficiency over DT. Although our algorithm’s navigation time is marginally longer than ODT's by 0.2 seconds, this difference arises from its comprehensive consideration of both the temporal and spatial states of surrounding pedestrians, facilitating a controlled deceleration upon approaching crowded areas. This deliberate deceleration contributes significantly to improved navigation safety.

\subsection{Real-world Experiment}
\begin{figure*}[htbp]
\centering
\subfloat[$t=0 \ s$]{
\includegraphics[trim=60 180 94 80, clip, scale=0.15]{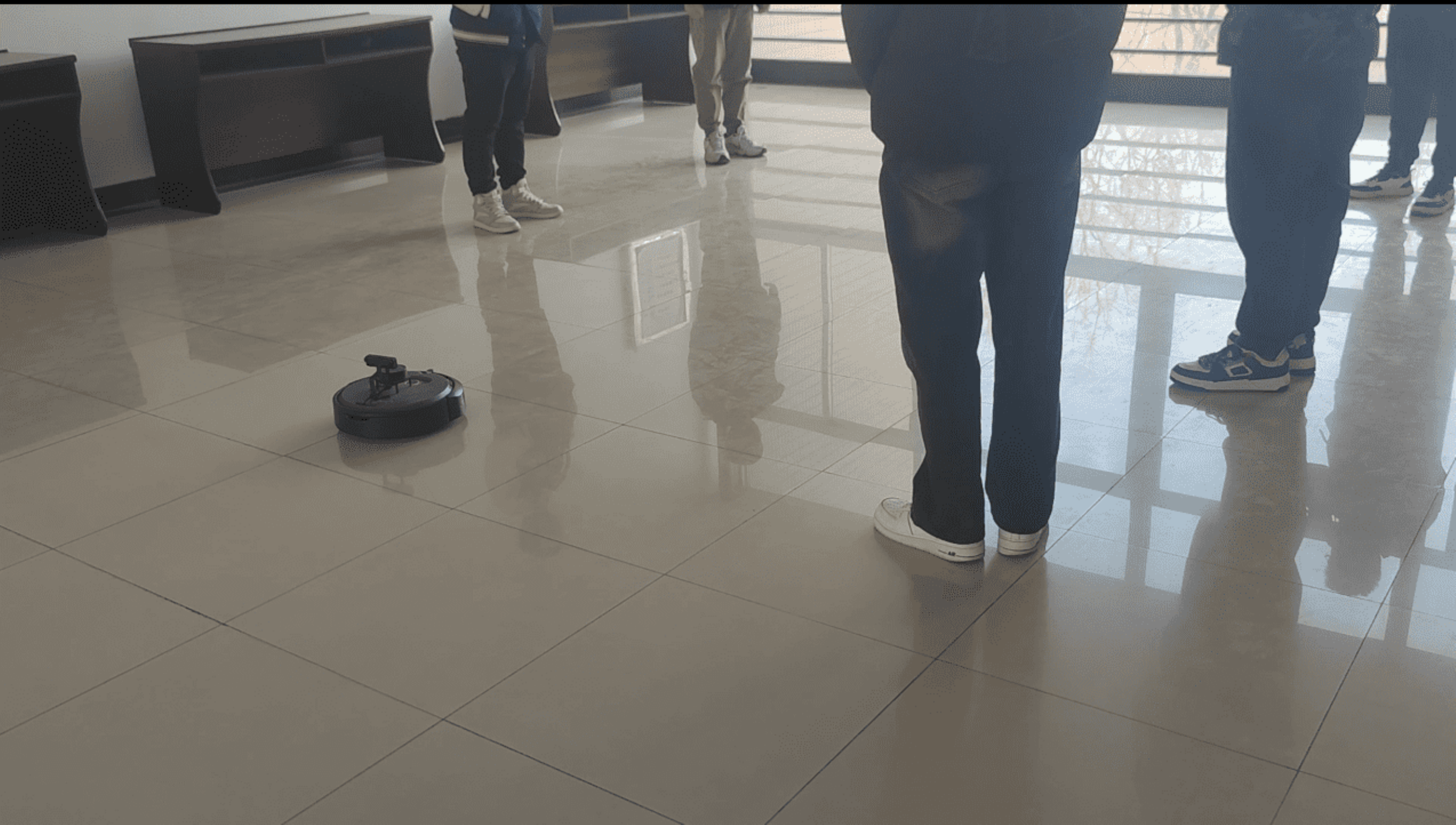}}
\subfloat[$t=11 \ s$]{
\includegraphics[trim=60 180 94 80, clip, scale=0.15]{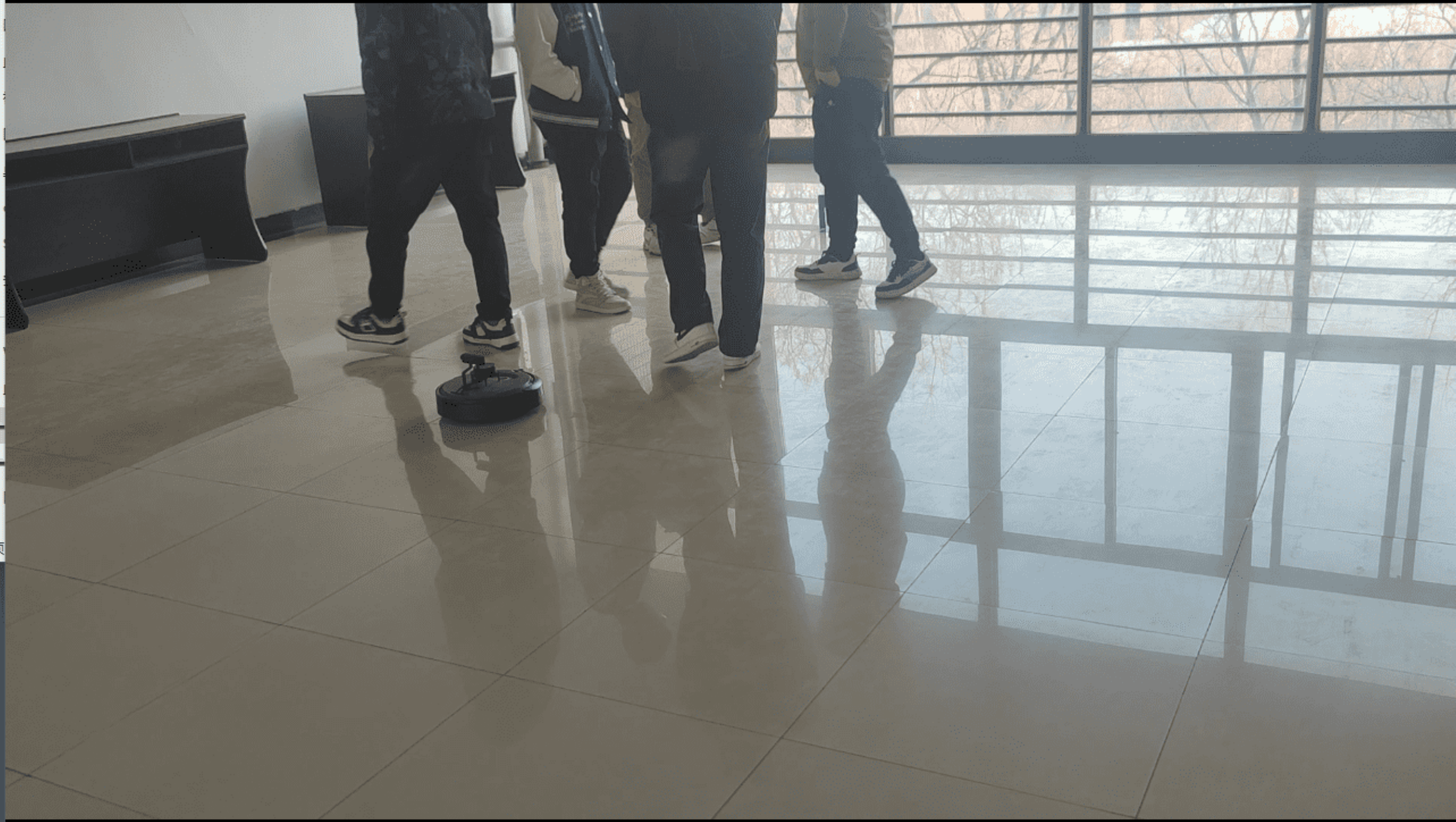}}
\subfloat[$t=25 \ s$]{
\includegraphics[trim=60 180 94 80, clip, scale=0.15]{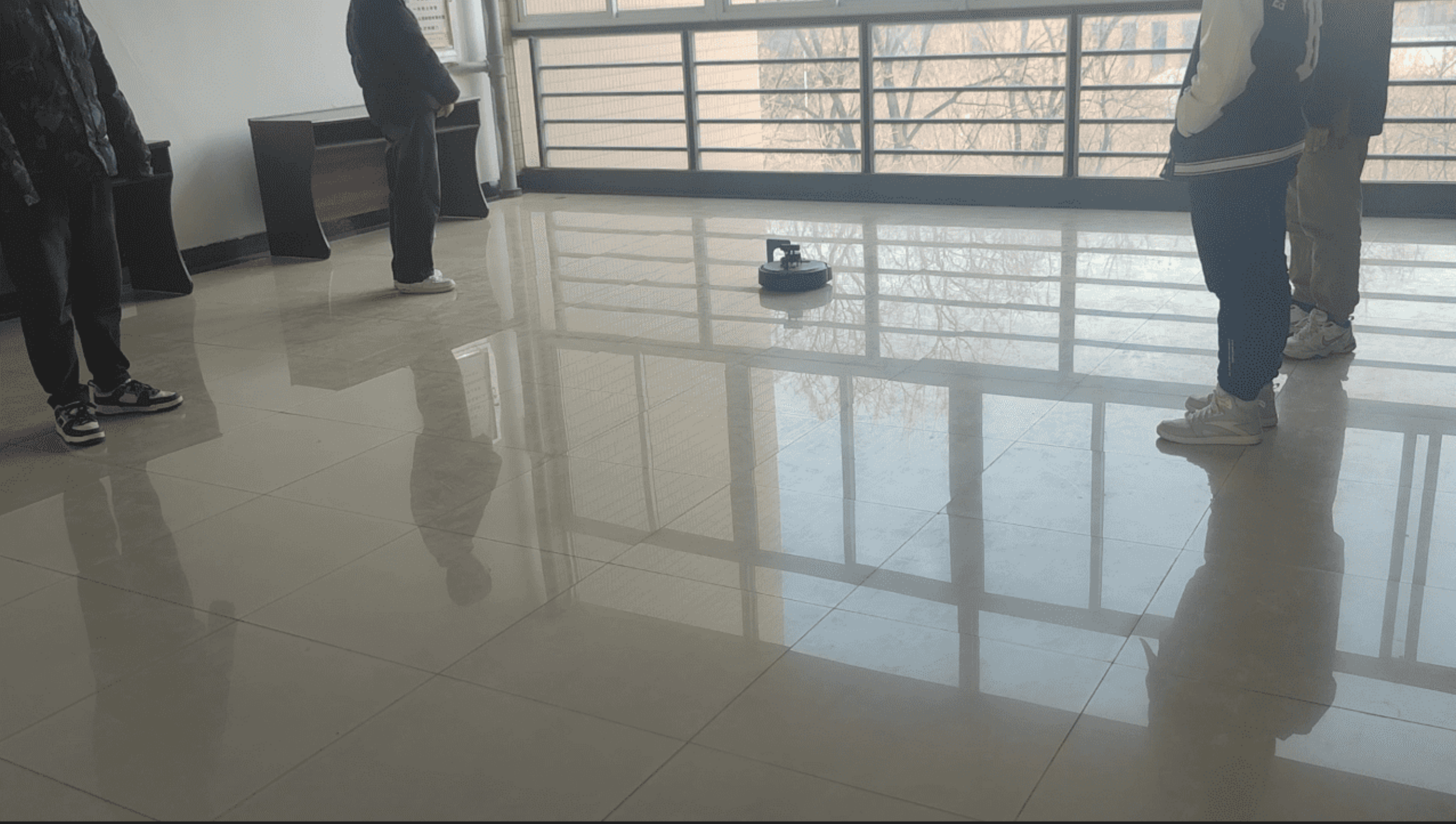}}
\caption{Testing of robots in real situations}
\label{Testing3}
\end{figure*}

We conducted real-world experiments, as shown in Figure ~\ref{Testing3}.  The robot is equipped with an RPLIDAR-A1 radar and employs a human leg detection algorithm\footnote{https://github.com/ShelyH/leg\_detector\_ros2} to estimate the speed and position of pedestrians.  The method runs on a laptop with an R9-7940HX CPU and an RTX4060 GPU.

\begin{figure}[htbp]
\centering
\includegraphics[scale=0.15]{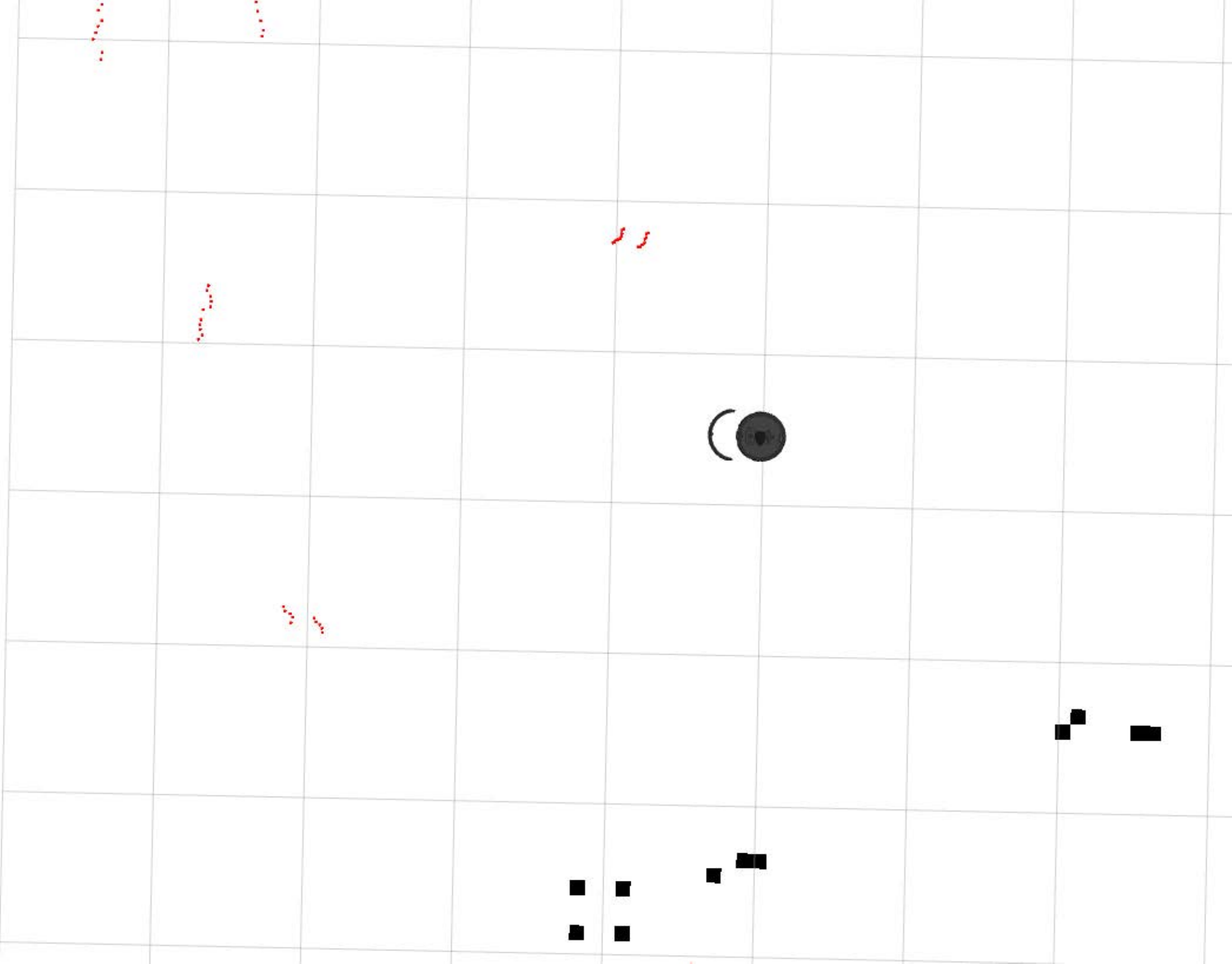}
\caption{The real-world radar map }
\label{map}
\end{figure}

The robot gathers data and sends it to the computer for processing, where the next action is determined based on the approach.  In parallel, we model the real-world environment to collect the data necessary for our approach.  The real-world radar map is shown in Figure ~\ref{map}.  In this map, the black dot represents the robot, and the grid layout aids in the robot's coordinate calculations.  The red dots represent the positions of pedestrians detected by the robot.  The robot’s state is monitored through its built-in chassis odometry sensor.

For each trial, we set the robot's target a few meters ahead of its starting position.  During navigation, pedestrians pass in front of the robot, simulating its obstacle avoidance behavior.  The robot successfully estimates the pedestrians' states and navigates to the target without colliding with any of the five pedestrians, as shown in Figure~\ref{map}.  A real-world demonstration of our method can be found in Video Material 1 in the appendix.  These results show that our approach effectively transfers from simulation to real-world robotic applications, ensuring a safe and reliable navigation strategy.

\section{Conclusion}
In this study, we have propose the OTOFRL algorithm to address distribution shift in social robot navigation caused by the complexity and unpredictability of pedestrian movements. By introducing a RTGP model and employing a hybrid offline-online sampling technique, OTOFRL ensures a seamless transition from offline pre-training to online fine-tuning, enhancing adaptability to new data in human-robot interaction environments. Experimental results show that our approach achieves state-of-the-art performance in success rate, sample efficiency, and average reward, outperforming existing methods in social navigation tasks.

\bibliographystyle{IEEEtran}
\bibliography{bibtex}

\vfill

\end{document}